\documentclass[letterpaper, 10 pt, conference]{ieeeconf}  

\IEEEoverridecommandlockouts                              

\usepackage{epsfig}
\usepackage{times}
\usepackage{amsmath}
\usepackage{amssymb}
\usepackage{graphicx}
\usepackage{cite}
\usepackage{subcaption}
\usepackage[]{algorithm2e}
\usepackage{framed}

\usepackage{multirow} 
\usepackage{booktabs} 

\title{\LARGE \bf
DESIGN OF THE FIRST SUB-MILLIGRAM FLAPPING WING AERIAL VEHICLE 
}

\author{\textit{Palak Bhushan and Claire Tomlin} \\ 
Dept. of EECS, University of California Berkeley, CA, USA 
}

\begin{document}

\maketitle
\thispagestyle{empty}
\pagestyle{empty}

\begin{abstract} 
Here we report the first sub-milligram flapping wing vehicle which is able to mimic insect wing kinematics. Wing stroke amplitude of 90$^\circ$ and wing pitch amplitude of 80$^\circ$ is demonstrated. This is also the smallest wing-span (single wing length of 3.5mm) device reported yet and is at the same mass-scale as a fruit fly. Assembly has been made simple and requires gluing together 5 components in contrast to higher part count and intensive assembly of other milligram-scale microrobots. This increases the fabrication speed and success-rate of the fully fabricated device. Low operational voltages (70mV) makes testing further easy and will enable eventual deployment of autonomous sub-milligram aerial vehicles. 
\end{abstract} 

\section*{\textbf{INTRODUCTION}}

Majority of milligram-scale flapping wing devices reported till date lie in the 100mg mass range \cite{wood_liftoff,wood_liftoff07,baybug18,robofly18}, with one weighing 3mg \cite{lin_electrostatic} but aimed as an actuator for 100mg-scale devices with $\approx$ 3cm wing spans. This is in part because to mimic insect wing kinematics one needs to produce large wing strokes. It is very tough to do so using other designs like the SCM based fabrication reported in \cite{wood_liftoff07} because they are already at $\approx 70\mu$m feature sizes to amplify small piezo displacements for 100mg-scale vehicles and going further down to accommodate for even smaller piezo motion is non-trivial. The work reported here is at 100$\mu$m feature sizes (excluding the wing) even at 1mg-scale. 

A 1mg vehicle has the same advantages over a 100mg device that a 100mg device has over a 10g device. 100 smaller robots can be used in place of one big robot. This multiplicity expectedly more than compensates for any deterioration in sensor quality and locomotion, while exponentially decreasing manufacturing cost per unit. 

The electromagnetic (EM) actuator and spring design are borrowed from \cite{baybug18}. This device has been scaled down 2 orders of magnitude in mass. It is low-voltage in operation just like its parent design. This is in contrast to most other milligram-scale aerial vehicles and microrobots that need 200-5000V \cite{wood_liftoff,robofly18,lin_electrostatic,inchworm12} to operate, and thus struggle with heavy and inefficient power electronics units to drive them \cite{robofly18}. 

Wing design including passive wing pitch is similar to that reported in \cite{wood_liftoff07,baybug18}, but we perform some modifications over this procedure to make wings lighter. In addition to the flapping wing device, these wings too are the lightest reported till date but are still heavier than wings of similar sized insects. 

Most of the subcomponents of this device are laser cut from a single material sheet just like its parent design. This, along with the low part count and higher feature sizes makes manual assembly very easy and has in part enabled the fabrication of this first sub-milligram flapping wing vehicle.

The milligram-scale aerial vehicles mimic insect wing kinematics to function, but in turn also provide insights and help study aerodynamics at small scales. The device presented here will enable, for the first time, an active study and exploration of flight at the fruit fly scale which is at a low Reynolds number of $\approx$ 100. 

\begin{figure}
\centering
\epsfig{file=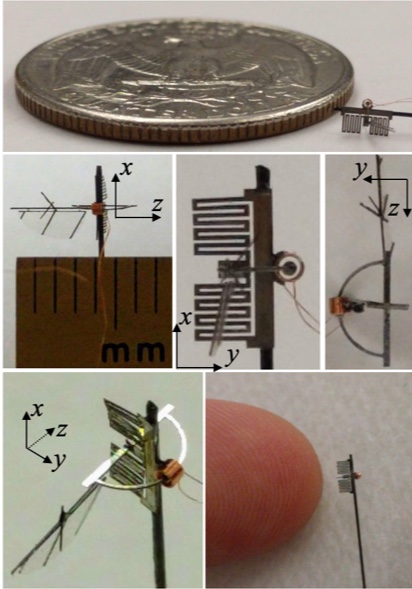,width=3.4in}
\vspace{-1.7em}
\caption{{\textit{\textbf{Assembled device.} (Top) Comparison with a quarter dollar coin. (Middle) Front, side and top views of the device. Front view is pictured with a millimeter ruler. (Bottom) Perspective view of the device, and comparison with an index finger.}} }
\vspace{-2em}
\label{fig:1}
\end{figure}

\section*{\textbf{METHODOLOGY}}

Here we describe the actuator and wing design, followed by the device assembly.

\subsection*{ELECTROMAGNETIC ACTUATOR} 
The actuation scheme uses Lorentz force produced in a magnet-coil system to produce mechanical power. Here we use a miniaturized version of the actuator reported in \cite{baybug18}. To directly produce large wing strokes the magnet is moved through the coil along a circular arc with large angular motion (see Fig. 2). Please see \cite{baybug18} to find the details of the actuator.

\begin{figure}[h]
\vspace{-0.6em}
\centering
\epsfig{file=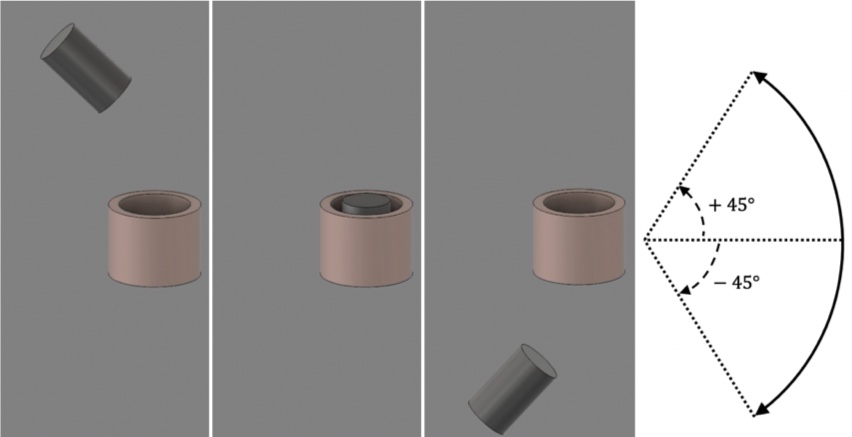,width=3.4in}
\vspace{-1.7em}
\caption{{\textit{\textbf{Magnet motion.} The desired circular arc the magnet should move in. The motion is simple harmonic in the magnet’s rotation angle with $\pm 45^\circ$ amplitude.}} }
\vspace{-1.0em}
\label{fig:2}
\end{figure}

\begin{figure}[h]
\vspace{-0.6em}
\centering
\epsfig{file=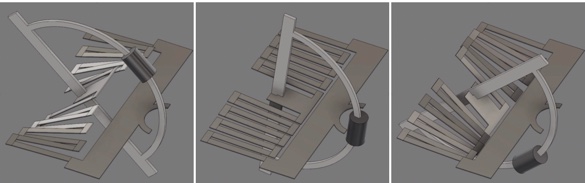,width=3.4in}
\vspace{-1.7em}
\caption{{\textit{\textbf{Spring motion.} The designed torsion spring in its extreme top, neutral, and extreme bottom positions. This shows the intended circular trajectory of the magnet. }}}
\vspace{0em}
\label{fig:3}
\end{figure}

The magnet used is Neodymium grade N52 with a height of 0.5mm and a diameter of 0.3mm. The coil is made out of 25$\mu$m Copper wire with $2 \times 14$ number of windings and is 0.45mm in height and has 0.45mm internal diameter. We set the radius of the arc the magnet should move in at $r=1.4$mm to provide sufficient clearance between the magnet and the coil. We use a torsion spring to restrict the motion of the magnet along the desired circular arc (see Figs. \ref{fig:2}, \ref{fig:3}). 

Fruit flies at similar size scales have wing stroke frequencies around $\approx$ 200Hz and wing mass around $\approx$ 5ug \cite{drosophila66}. However, the lightest wings we can currently manufacture weigh 4$\times$ times as much (see Table \ref{table:mass-dist}). Thus, our wing resonance frequency will be approximately half that of the fruit fly. In order to operate the wing quasi-statically we need wing stroke frequency $\ll$ wing resonance frequency \cite{Di02,Di99,passive_rot}, and so we scale down the wing stroke frequency by a factor of 2$\times$ to be near 100Hz. 

In order for the magnet-spring system to have a target resonance frequency of, say, $f=130$Hz, the torsional stiffness of the spring should be $m_{magnet} r^2 (2\pi f)^2= 0.34\mu$Nm. To take into account additional inertia of the glue and frames we choose spring stiffness to be 0.8$\mu$Nm to be on the safe side. We can always tune down the resonance frequency post-fabrication by adding more mass via glue. 

Torsion spring of the desired stiffness is fabricated using the procedure outlined in \cite{baybug18}. The material used here is a 12.7$\mu$m-thick stainless-steel sheet which is laser cut to make the planar spring. The dimensions of the spring are optimized using 3D FEA simulations such that there is negligible parasitic off-axes motions and resonances. Resulting spring dimensions are reported in Table \ref{table:spring-spec}.

\begin{table}[h]
\vspace{-0.4em}
\normalsize
\centering
\caption{\label{table:spring-spec}\textit{\textbf{Spring specifications.}}}
\vspace{-0.6em}
\begin{tabular}{|c|c|}
  \hline
 \# parallel beams & 16 \\
 \hline
  Length of each beam & 1mm \\
  \hline
  Beam width & 0.1mm \\
  \hline
  Beam thickness & 12.7$\mu$m \\
  \hline
\end{tabular}
\vspace{-0.4em}
\end{table}

\subsection*{WING FABRICATION} 
Wing design is similar to that reported in \cite{wood_liftoff07,baybug18} with flexures included for passive wing pitch. However, there is one key difference. In order to minimize the wing’s rotational inertia (to maximize wing resonance frequency), the veins are made from a single layer of 30$\mu$m-thick unidirectional carbon fiber (uni-CF) sheet (30$\mu$m was the thinnest CF prepreg sheet we could obtain), as opposed to from a thicker sheet with multiple cured layers with each layer’s fibers running along different directions. Due to this design choice we need to ensure that fibers always run along the vein direction in order to strengthen it since uni-CF is weak along the transverse direction. 

\begin{figure}[h]
\centering
\epsfig{file=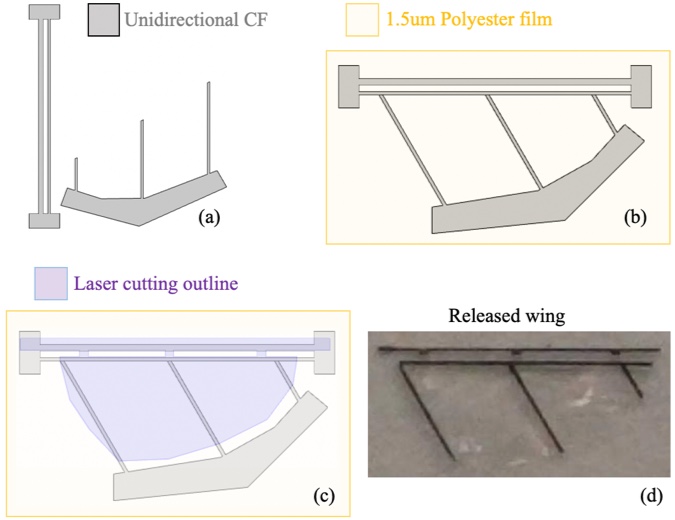,width=3.4in}
\vspace{-1.7em}
\caption{{\textit{\textbf{Steps of wing fabrication.} (a) CF veins are laser cut from a unidirectional single layer 30$\mu$m-thick CF sheet. The fibers are oriented vertically. (b) Laser cut veins are aligned and adhered to a polyester film using 18$\mu$m-thick adhesive layer. (c) The resulting sandwich is laser cut to remove the excess CF and to form the flexures. (d) Released wing. }}}
\vspace{-1em}
\label{fig:4}
\end{figure}

A 18$\mu$m-thick adhesive sheet is first bonded to the uni-CF sheet. We then laser cut the uni-CF such that the leading edge and all the veins are 30$\mu$m wide and aligned in the same direction (see Fig. \ref{fig:4}(a)). 30$\mu$m was found to be the narrowest beam we could cut using our UV laser cutter. The veins are placed in the final orientation and then bonded to a 1.5$\mu$m polyester membrane using the previously applied adhesive layer (see Fig. \ref{fig:4}(b)). This assembly is then laser cut to form flexures along the leading edge from the same membrane material (see Fig. \ref{fig:4}(c)), and the wing is released. The wing length is chosen to be 3.5mm in order to be of similar size to similar sized insects \cite{drosophila66}. The shape is chosen for the wing to have an aspect ratio of $\approx$ 3. 

For the wings to deflect by a maximum of 1rad $\approx 60^\circ$, the flexure stiffness should be the same as the maximum aerodynamic torque experienced by the wing along the leading edge. Assuming the center of pressure to be 0.4mm away from the leading edge, and the maximum normal force seen by a single wing to be = $0.5\cdot \sqrt{2} \cdot ($average lift = 0.01mN) = 0.007mN, the maximum aerodynamic torque is estimated at 0.0028$\mu$Nm. For a $w$-wide and $l$-long flexure of $t$ = 1.5$\mu$m thick polyester membrane with an elastic modulus of $E$ = 2.5GPa, the stiffness is given by $\frac{E}{12} t^3 \frac{w}{l}$. This gives one possible desired flexure width = 390$\mu$m and flexure length = 100$\mu$m. To eliminate any off-axis twisting torques that may be caused by the aerodynamic loading, the flexure is made in 3 parts each 130$\mu$m wide and spread out throughout the leading edge of the wing (see Fig. \ref{fig:4}(c)), thus ensuring the flexure only bends along a single axis.  

\subsection*{DEVICE ASSEMBLY}

\begin{figure}[h]
\centering
\epsfig{file=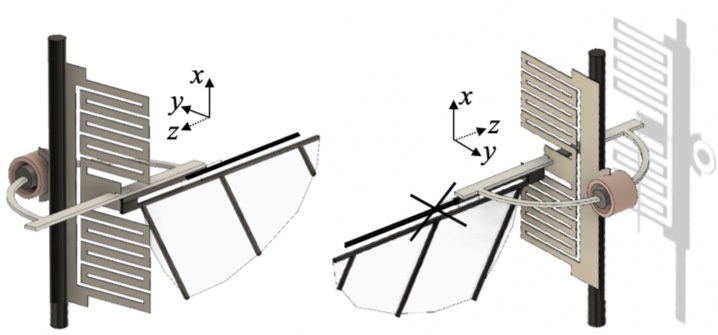,width=3.4in}
\vspace{-1.7em}
\caption{{\textit{\textbf{Assembled device, animation.} Axes defined with respect to the assembled body. The shadow shows the concentricity of the coil and the magnet, and the clearance between them. }}}
\vspace{-1.2em}
\label{fig:5}
\end{figure}

See Fig. \ref{fig:5}. The planar steel spring in the xy-plane is glued to a 0.28mm diameter CF rod for eventual ease of device handling and mounting. The steel spring has a curved slot in it to position the coil. A planar D-shaped 50$\mu$m-thick laser-cut Aluminum frame in the yz-plane is glued perpendicularly to the spring. This D-frame has a gap in the middle of its curved part to insert the magnet. The coil and the magnet with their axis along z are glued to the spring and frame, respectively, so that they are concentric while in spring’s neutral position. The wing in the xz-plane is glued to the straight part of the D-frame at the flexure’s top supporting edge. A thin X-shaped CF frame is glued on to the flexure’s top supporting edge to limit the wing pitch amplitude. The segments of the X-frame collide with the central wing vein when the wing plane approaches a certain pitch value in either direction. This stops the flexure and wing plane from rotating any further. The coil is connected to a standard function generator. The mass distribution of the assembled device is reported in Table \ref{table:mass-dist}. 

\begin{table}[h]
\normalsize
\centering
\caption{\label{table:mass-dist}\textit{\textbf{Mass distribution.}}}
\vspace{-0.6em}
\begin{tabular}{|c|c|}
  \hline
  Coil & 0.25mg \\
 \hline
  Magnet & 0.26mg \\
  \hline
  Spring & 0.15mg \\
  \hline
  D-frame & 0.05mg \\
  \hline
  Wing & 0.02mg \\
  \hline
  \hline 
  Net & 0.7mg \\
  \hline
\end{tabular}
\vspace{-0.8em}
\end{table}

\section*{\textbf{RESULTS}}
\begin{figure}
\centering
\epsfig{file=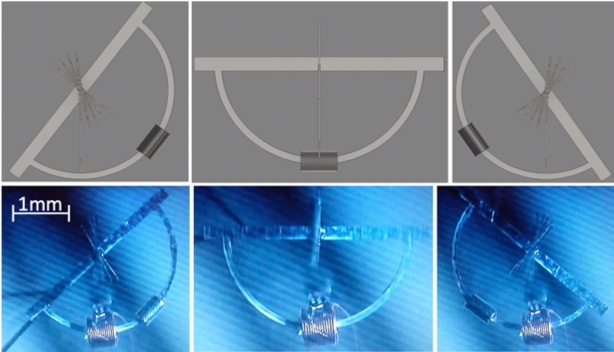,width=3.4in}
\vspace{-1.7em}
\caption{{\textit{\textbf{Magnet motion snapshots, top view.} Extreme right, neutral, and extreme left positions of the moving magnet plus spring system. (Top) An animation of magnet and spring positions. (Bottom) Snapshots of the fabricated device in motion, with Copper coil being stationary.}}}
\vspace{-0.4em}
\label{fig:6}
\end{figure}

\begin{figure}
\centering
\epsfig{file=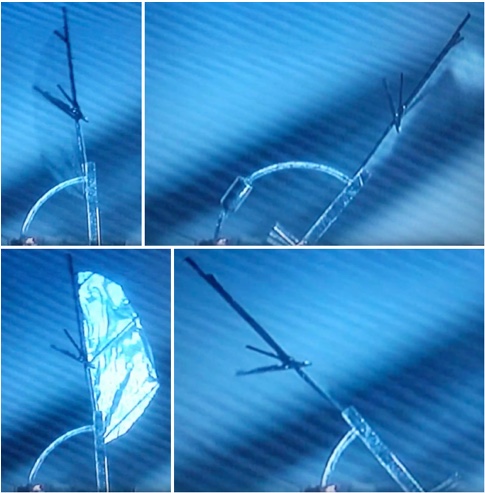,width=3.4in}
\vspace{-1.7em}
\caption{{\textit{\textbf{Wing pitch, top view.} (Top) Positive wing pitch (that is, positive angle of attack) while moving to the right. A maximum pitch of $30^\circ$ is observed. Zero pitch observed at extreme stroke angle. (Bottom) Wing pitch reversed while moving to the left. Maximum pitch of $50^\circ$ observed. Zero pitch at extreme stroke angle. }}}
\vspace{-1em}
\label{fig:7}
\end{figure}

\begin{figure}
\centering
\epsfig{file=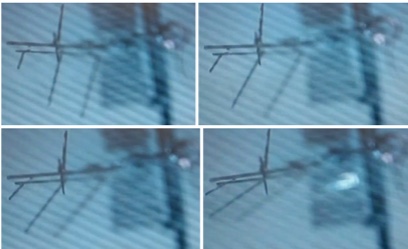,width=3.4in}
\vspace{-1.7em}
\caption{{\textit{\textbf{Wing pitch, side view.} Wing pitch reversal at the end of a stroke cycle (that is, near an extreme stroke angle). The X-shaped CF frame stops the wing from pitching further after reaching a certain angle of attack. This limiting can be seen more clearly in Fig. \ref{fig:9}.  }}}
\vspace{-0.4em}
\label{fig:8}
\end{figure}

\begin{figure}
\centering
\epsfig{file=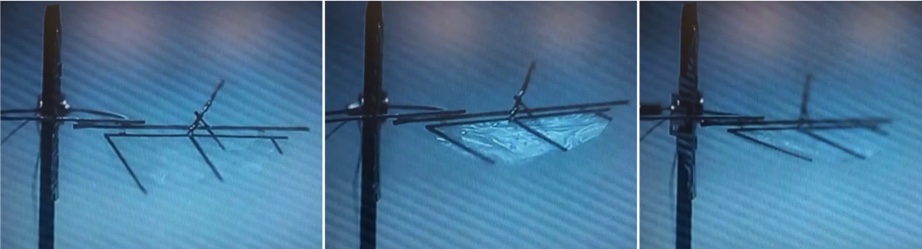,width=3.4in}
\vspace{-1.7em}
\caption{{\textit{\textbf{Wing pitch, front view.} Wing pitch amplitude increases as the mechanism’s stroke speed increases. The X-shaped CF frame hard-limits the pitch magnitude. }}}
\vspace{-1em}
\label{fig:9}
\end{figure}

The coil is driven by a square wave and the motion of the device is observed using strobe lights under a microscope. For simplicity, and to reduce the number of steps in the assembly, only one wing is attached to the actuator. Resonance is observed at 132.3Hz, and a $\pm 45^\circ$ wing stroke is achieved with a $\pm$70mV applied square wave voltage (see Fig. \ref{fig:6}). A wing pitch of $+30^\circ$/$-50^\circ$ is observed with pitch magnitude maximums at neutral stroke angle and zero pitch at extreme stroke angles (see Fig. \ref{fig:7}). The asymmetry in wing pitch is due to manual assembly imperfections like the wing plane not being perfectly in the xz-plane and the placement of the X-frame. Wing pitch reversal can be observed at extreme stroke angles (see Fig. \ref{fig:8}). The X-frame can be seen in action when the wing plane tries to pitch more than the set limit (see Fig. \ref{fig:9}). 

The resistance of the coil is $\approx 1.5\Omega$ meaning Joule heat loss is $\approx (0.07)^2\cdot 1.5=3.3$mW. Fruit flies have a body-mass-specific power of $\approx 29$W/kg \cite{drosophila02} meaning that for producing $\approx 1$mg of lift a mechanical power of $\approx 29\mu$W is required. We noticed in \cite{baybug18} that since the wing shape and trajectory aren’t optimized, the lift generated is about 60\% the designed value and the mechanical power consumed is 1.6 times than was theoretically needed to generate the designed lift. We expect a similar behavior here since this work is a miniaturized version of \cite{baybug18}. Thus, with a single wing, we expect a mechanical power output of 23$\mu$W generating 0.3mg of lift. Presently we lacked the capacity to measure $\approx 0.1$mg lift forces. The above figures give the estimated electromechanical efficiency of our device as 0.7\%. 

\section*{\textbf{CONCLUSION}}
In this paper we constructed a device that is 2 orders of magnitude lighter than all other flapping wing devices reported till date, and which is able to mimic insect wing kinematics. We estimated the lift generated by the device, but a sensitive anemometer can be used to precisely measure the lift in future work. Current battery technology and power electronics aren’t ready for even 100mg-scale devices so autonomous flight for 1mg-scale devices will have to wait. The efficiency of fruit fly muscles is $\approx 17$\% which is an order of magnitude higher than our actuator \cite{drosophila02}. Meanwhile, we can put our efforts in developing more efficient actuators to be ready for newer batteries and power electronics units, and also design appropriate sub-100$\mu$g sensors and controllers for these devices.

\section*{\textbf{ACKNOWLEDGEMENTS}}

The authors are grateful to get support from Commission on Higher Education (award \#IIID-2016-005)
and DOD ONR Office of Naval Research (award \#N00014-16-1-2206). 
We would also like to thank Prof. Ronald Fearing for his help and insightful discussions.

\section*{\textbf{CONTACTS}} 
$\{$palak,tomlin$\}$@berkeley.edu


\begin{thebibliography}{99}

\bibitem{wood_liftoff} K. Ma, P. Chirarattanon, S. Fuller, and R.J. Wood, ``Controlled Flight of a Biologically Inspired, Insect-Scale Robot,'' {Science}, vol. 340, pp. 603-607, 2013. 

\bibitem{wood_liftoff07} R. J. Wood, ``Liftoff of a 60mg flapping-wing MAV,'' IROS, San Diego, CA, Oct. 2007. 

\bibitem{baybug18} P. Bhushan and C.J. Tomlin, ``Milligram-scale Micro Aerial Vehicle Design for Low-voltage Operation,'' {IROS}, Madrid, Spain, Oct. 2018. 

\bibitem{robofly18} J. James, V. Iyer, Y. Chukewad, S. Gollakota, and S.B. Fuller, ``'Liftoff of a 190 mg Laser-Powered Aerial Vehicle: The Lightest Untethered Robot to Fly,'' {IEEE Int. Conf. on Robotics and Automation}, Brisbane, Australia, May 2018. 


\bibitem{lin_electrostatic} X. Yan, M. Qi, and L. Lin, ``Self-Lifting Artificial Insect Wings via Electrostatic Flapping Actuators,'' {Proceedings of 28th IEEE Micro Electro Mechanical Systems Conference}, pp. 22-25, Portugal, Jan. 2015.

\bibitem{inchworm12} I. Penskiy and S. Bergbreiter, ``Optimized electrostatic inchworm motors using a flexible driving arm,'' {Journal of Micromechanics and Microengineering}, Vol. 23, No. 1, pp 1-12, 2012.

\bibitem{drosophila66} S. Vogel, ``Flight in Drosophila. I. Flight Performance of Tethered Flies,'' The Jour. Of Exp. Biol., 44: 567-578, 1966.

\bibitem{Di02} S. P. Sane and M. H. Dickinson, ``The aerodynamic effects of wing rotation and a revised quasi-steady model of flapping flight,'' The Jour. of Exp. Biol., 205, 1087-1096, 2002.

\bibitem{Di99} M. H. Dickinson, F.-O. Lehmann, S. P. Sane, ``Wing rotation and the aerodynamics basis of insect flight,'' Science, vol. 284, pp. 1881-2044, 1999.

\bibitem{passive_rot} J. P. Whitney and R. J. Wood, ``Aeromechanics of passive rotation in flapping flight,'' J. Fluid Mech., vol. 660, pp. 197-220, 2010.

\bibitem{drosophila02} M. Sun, J. Tang, ``Lift and power requirements of hovering flight in Drosophila virilis,'' The Jour. Of Exp. Biol., 205: 2413-2427, 2002.

\end{thebibliography}
\end{document}